\documentclass{article}
\usepackage{ijcai25}

\usepackage{times}
\usepackage{soul}
\usepackage{url}
\usepackage[hidelinks]{hyperref}
\usepackage[utf8]{inputenc}
\usepackage[small]{caption}
\usepackage{graphicx}
\usepackage{amsmath}
\usepackage{amsthm}
\usepackage{booktabs}
\usepackage{algorithm}
\usepackage{algorithmic}
\usepackage[switch]{lineno}

\usepackage{multirow}
\usepackage{booktabs}
\usepackage{graphicx}
\usepackage{amssymb} 
\usepackage[table]{xcolor}

\usepackage{tabularx}

\title{Visual Perturbation and Adaptive Hard Negative Contrastive Learning for Compositional Reasoning in Vision-Language Models}

\author{
    Author Name
    \affiliations
    Affiliation
    \emails
    email@example.com
}

\author{
Xin Huang$^{1,3}$\and
Ruibin Li$^{1,3}$\and
Tong Jia$^2$\and
Wei Zheng$^{1,3}$\and
Ya Wang$^{1,3*}$\\ 
\affiliations
$^1$School of Artificial Intelligence and Software Engineering, Nanyang Normal University, Henan, China\\
$^2$Institute for Artificial Intelligence, Peking University, Beijing, China\\
$^3$Collaborative Innovation Center of Intelligent Explosion-proof Equipment, Henan, China\\
\emails
huangxin@nynu.edu.cn,
liruibin199810@nynu.edu.cn,
jia.tong@pku.edu.cn,
zhengwei821@nynu.edu.cn,
wangya@nynu.edu.cn
}

\begin{document}

\maketitle

\begin{abstract}

Vision-Language Models (VLMs) are essential for multimodal tasks, especially compositional reasoning (CR) tasks, which require distinguishing fine-grained semantic differences between visual and textual embeddings. However, existing methods primarily fine-tune the model by generating text-based hard negative samples, neglecting the importance of image-based negative samples, which results in insufficient training of the image encoder and ultimately impacts the overall performance of the model. Moreover, negative samples are typically treated uniformly, without considering their difficulty levels, and the alignment of positive samples is insufficient, which leads to challenges in aligning difficult sample pairs. To address these issues, we propose Adaptive Hard Negative Perturbation Learning (AHNPL). AHNPL translates text-based hard negatives into the visual domain to generate semantically disturbed image-based negatives for training the model, thereby enhancing its overall performance. AHNPL also introduces a contrastive learning approach using a multimodal hard negative loss to improve the model's discrimination of hard negatives within each modality and a dynamic margin loss that adjusts the contrastive margin according to sample difficulty to enhance the distinction of challenging sample pairs. Experiments on three public datasets demonstrate that our method effectively boosts VLMs' performance on complex CR tasks. The source code is available at \url{https://github.com/nynu-BDAI/AHNPL}.

\end{abstract}

\section{Introduction}

In recent years, Vision-Language Models (VLMs) have achieved significant zero-shot performance advancements in multimodal tasks such as retrieval \cite{Huang2,APSE-IPIK}, classification \cite{AutoCLIP,CHILS} and segmentation \cite{GroupViT,CLIPSeg}, establishing themselves as a foundation for multimodal tasks. However, despite their impressive performance, VLMs still face notable challenges in handling compositional reasoning (CR) tasks \cite{DAC,Winoground}. CR tasks demand that models accurately understand and reason about fine-grained relationships among objects, attributes, and actions in images or texts. For instance, they must be able to distinguish sentences differing only in word order or accurately identify objects in images that have similar colors but different attributes. Such tasks go beyond the need for coarse-grained semantic understanding, requiring precise reasoning in complex scenarios.

\begin{figure}
    \centering
    \includegraphics[width=1\linewidth]{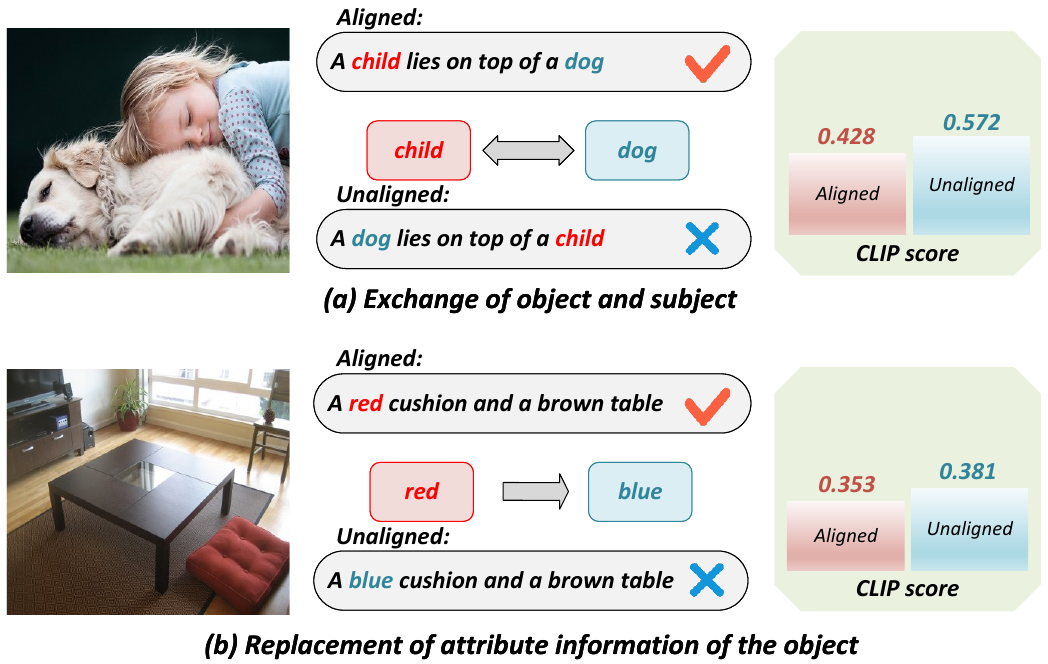}
    \caption{CLIP scores of the image and corresponding descriptions. The “Aligned” descriptions correctly reflect the image content, while the “Unaligned” descriptions represent incorrect descriptions of the image. }
    \label{fig:1}
\end{figure}

Yuksekgonul et al. \cite{Neg-CLIP} highlight the limitations of existing VLMs in CR tasks. They found that current state-of-the-art VLMs often perform poorly in distinguishing semantic differences caused by subtle changes in word order or attribute variations. As demonstrated in Fig.~\ref{fig:1}, the CLIP scores show that, in some cases, the unaligned descriptions receive higher scores, indicating that VLMs still struggle to distinguish fine-grained semantic differences in object relationships and attributes. To address this issue, researchers attempted to enhance models’ ability to discern complex semantic relationships by generating hard negative samples. These hard negative samples are semantically similar to the positive samples but have subtle semantic differences. By generating these challenging samples to fine-tune the model, its performance on CR tasks is improved.

However, current methods for incorporating hard negative samples into contrastive learning face several issues. First, some methods focus solely on hard negatives in the text modality, overlooking equally important hard negatives in the visual modality \cite{Structure}, resulting in insufficient training of the image encoder and thereby affecting the overall model performance. Second, these methods typically treat all negative samples equally, without considering their varying difficulty levels \cite{MSKR,CounterCurate}. This prevents the model from effectively capturing subtle differences in hard negatives during training, thus hindering its ability to learn fine-grained distinctions. Additionally, these methods often neglect the adjustment of alignment for positive samples, which may cause the model to align only simple positive pairs while struggling to align more challenging ones.

To address these issues, we propose Adaptive Hard Negative Perturbation Learning (AHNPL) to enhance VLMs' performance in CR tasks. Our AHNPL method explicitly maps the subtle semantic variations of text-based hard negatives into the visual space, generating corresponding image embeddings for these negative texts, thereby improving the model’s ability to discern subtle differences in the visual modality, which ultimately enhances the overall performance of the model. Moreover, AHNPL introduces dynamic hard negative contrastive learning, comprising a multimodal hard negative loss and a dynamic margin contrastive loss. The multimodal hard negative loss reduces the similarity between negative and positive samples across both text and visual modalities, improving the model’s ability to distinguish hard negatives. The dynamic margin contrastive loss adjusts the contrastive margin based on sample complexity, enabling the model to focus more on challenging negatives and positives, which improves the alignment quality of challenging sample pairs.

\noindent
\hspace{1em}Overall, our contributions are as follows:
\begin{itemize}
    \item We propose a novel method that generates image-based hard negative samples by mapping subtle semantic shifts from text-based negatives into the visual domain, fine-tuning the model to enhance its overall performance.
    \item We introduce a dynamic hard negative contrastive learning approach. This includes a multimodal hard negative loss to distinguish hard negatives across text and image, and a dynamic margin loss that adapts to sample difficulty, focusing on challenging examples.
\end{itemize}

\section{Related Work}
\paragraph{Contrastive Learning.} \hspace*{0.2cm} The objective of contrastive representation learning is to learn representations that are close to each other for similar samples and distant from each other for dissimilar samples. As one of the representative examples, CLIP \cite{CLIP} has become a milestone in this field. CLIP is a Transformer-based \cite{Transformer} model that consists of an image encoder and a text encoder, which are trained simultaneously. The objective is to maximize the cosine similarity of the image and text embeddings from the correct image-text pairs and to minimize the similarity between the incorrect pairs. A batch of N training samples (i.e., matching image-text pairs) results in a similarity matrix for each image-text combination. The main diagonal indicates the correct pair matches; the remaining entries correspond to negative entries. The InfoNCE loss is applied to the N × N similarity score matrix, and through contrastive learning, the image and text encoders are more effectively aligned across the two modalities, thus enhancing the performance of VLMs in multimodal tasks.

\begin{figure*}[ht] 
    \centering
    \includegraphics[width=\textwidth]{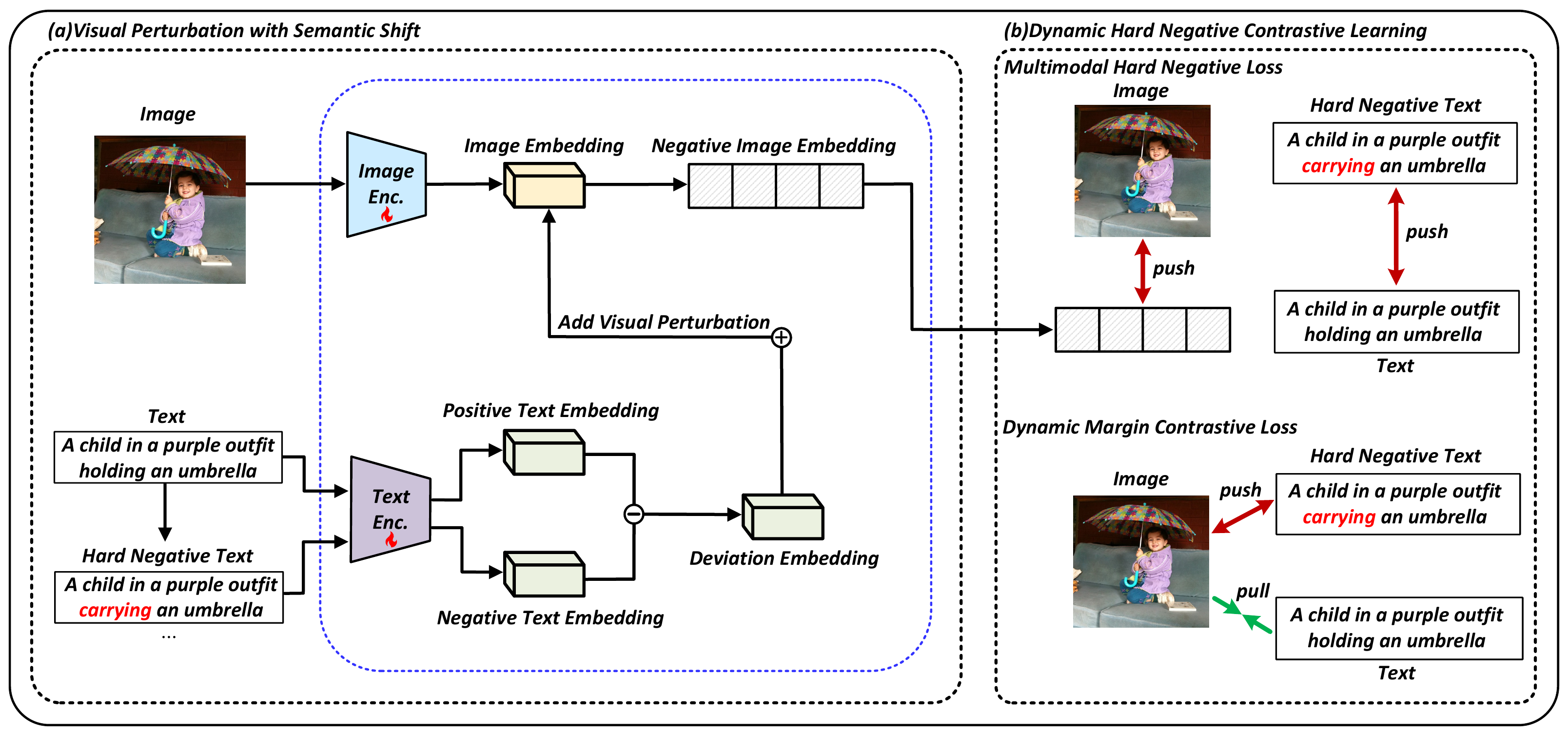}
    \caption{The proposed Adaptive Hard Negative Perturbation Learning (AHNPL). (a) Performing visual perturbation by computing the semantic shift in text to generate a deviation embedding. (b) The dynamic hard negative contrastive learning method makes targeted adjustments to the strategies for alignment of positive samples and the distinction of hard negative samples.}
    \label{fig:2}
\end{figure*}

\paragraph{Compositional Reasoning.} \hspace*{0.2cm}CR in vision and language tasks evaluates a model's ability to understand and manipulate complex ideas by breaking them into simpler components and recombining them in new ways. While VLMs excel at handling multimodal data, research shows they struggle with analyzing complex information, such as object relationships. Thrush et al. \cite{Winoground} first identified this issue, demonstrating that VLMs often fail to distinguish semantic differences caused by changes in word order, with performance sometimes no better than random guessing. This type of task, which tests a model's ability to understand fine-grained semantic structures, is referred to as CR. Many researchers have proposed methods of generating negative samples to fine-tune models to improve their performance in CR tasks. For example, Doveh et al. \cite{DAC} introduced DAC, which fine-tunes the model to improve its performance in CR tasks by using a Large Language Model (LLM) to generate texts of different scenes. Zhang et al. \cite{CE-CLIP} proposed CE-CLIP, which refines contrastive objectives with diverse negative samples to enhance semantic understanding. While effective, we believe current methods haven't fully explored hard negative mining, as they overlook the importance of image-based negative samples and fail to adjust the alignment process of positive sample pairs. To address this issue, we propose AHNPL, which translates text-based hard negatives into the visual domain, generating semantically disturbed image-based negatives, and dynamically adjusts the training process based on the similarity of each positive sample pair.

\section{Methodology}
In this section, we provide a detailed introduction to our proposed AHNPL method. Fig.~\ref{fig:2} shows an overview of the entire pipeline.
\subsection{Hard Negative Generation}
In contrastive learning, hard negatives are samples that are semantically similar to positive samples but have subtle semantic differences. For example, consider the caption: “A boy wearing a red hat is playing on the beach”. A potential hard negative could be: “A boy wearing a blue hat is playing on the beach”. While this hard negative accurately describes most elements in the image, it differs from the positive sample in the color of the hat. Including such hard negatives in the training process can help the model recognize subtle differences, thereby improving overall accuracy and performance.

To generate these hard negatives, we use the natural language processing tool Spacy \cite{spaCy} to parse the captions and assign part-of-speech tags to each word. Specifically, we generate two types of hard negatives. The first type involves swapping two nouns in a sentence to generate hard negatives. This aims to alter the relationships between entities, and thus train the model to learn how to distinguish between different entity relationships. The second type involves randomly masking words in the sentence based on their part-of-speech tags (such as nouns, verbs, or adjectives), and then using the RoBERTa \cite{RoBERTa} model to predict and fill in the masked parts. This approach aims to generate hard negatives that are similar in context and sentence structure, which helps the model enhance its sensitivity to semantic changes and lexical variations.

\subsection{Visual Perturbation with Semantic Shift}
In the previous section, we identify keywords in the captions through part-of-speech parsing and construct textual negative samples. Furthermore, we design a method for generating visual negative samples to train the model's image encoder, thereby enhancing the model's overall performance.

Existing negative sample generation methods often overly focus on textual negative samples, neglecting the role of visual negative samples, which leads to insufficient training of the image encoder and subsequently affects the overall model performance. To solve this problem, we propose a visual perturbation with semantic shift method for generating image negative samples. Specifically, we first capture semantic changes by computing the deviation embedding between the original text and the negative text. Given the original text \( T_{{orig}} \) and the negative text \( T_{{neg}} \), we use a pretrained CLIP model to extract their high-dimensional embedding vectors \( e_{T_{{orig}}} \) and \( e_{T_{{neg}}} \), respectively. These vectors represent the semantic information of the texts within the multimodal embedding space of the CLIP model. The deviation embedding \( \Delta_e \) is computed as follows:
\begin{equation}
\Delta_e  = e_{T_{{neg}}} - e_{T_{{orig}}}
\end{equation}
where \( \Delta_e \) reflects the semantic shift from the positive caption to the negative caption in the text. This shift not only represents changes in specific words or phrases within the text, but also implies adjustments in the contextual semantic structure. By utilizing this deviation embedding, we can capture the subtle semantic changes in the text.

Next, we directly incorporate the generated deviation embedding \( \Delta_e \) into the image embedding space by adding it to the original image's embedding vector \( e_{I_{{orig}}} \) to create the embedding representation \( e_{I_{{neg}}} \) for the image negative sample:
\begin{equation}
e_{I_{{neg}}} = e_{I_{{orig}}} + \Delta_e
\end{equation}

This approach ensures that the generated negative samples maintain appropriate semantic relevance to the original image in the embedding space, while deviating from the original image in terms of semantics. The image negative samples generated by this method maintain semantic consistency with the textual negative samples, providing more challenging training examples for the model.

\subsection{Dynamic Hard Negative Contrastive Learning}
Existing contrastive learning methods have some obvious limitations. First, these methods fail to adapt their alignment strategy according to the varying difficulties of positive samples, specifically by failing to adaptively enhance the learning signal for hard positive pairs that exhibit lower similarity, which causes the model to overly focus on aligning simple positive samples while lacking the capability to handle more challenging positive pairs. Additionally, these methods typically not only treat all negative samples as equivalent, disregarding the difficulty levels of negative samples during training, but also fail to effectively leverage multimodal information to distinguish the fine-grained semantic differences between hard negative samples and their corresponding positive samples. This approach makes it difficult for the model to establish clear decision boundaries and leads to misjudgments when processing hard negative samples that are semantically highly similar to positive samples, thereby affecting the effectiveness of contrastive learning. 

\begin{figure}
    \centering
    \includegraphics[width=1\linewidth]{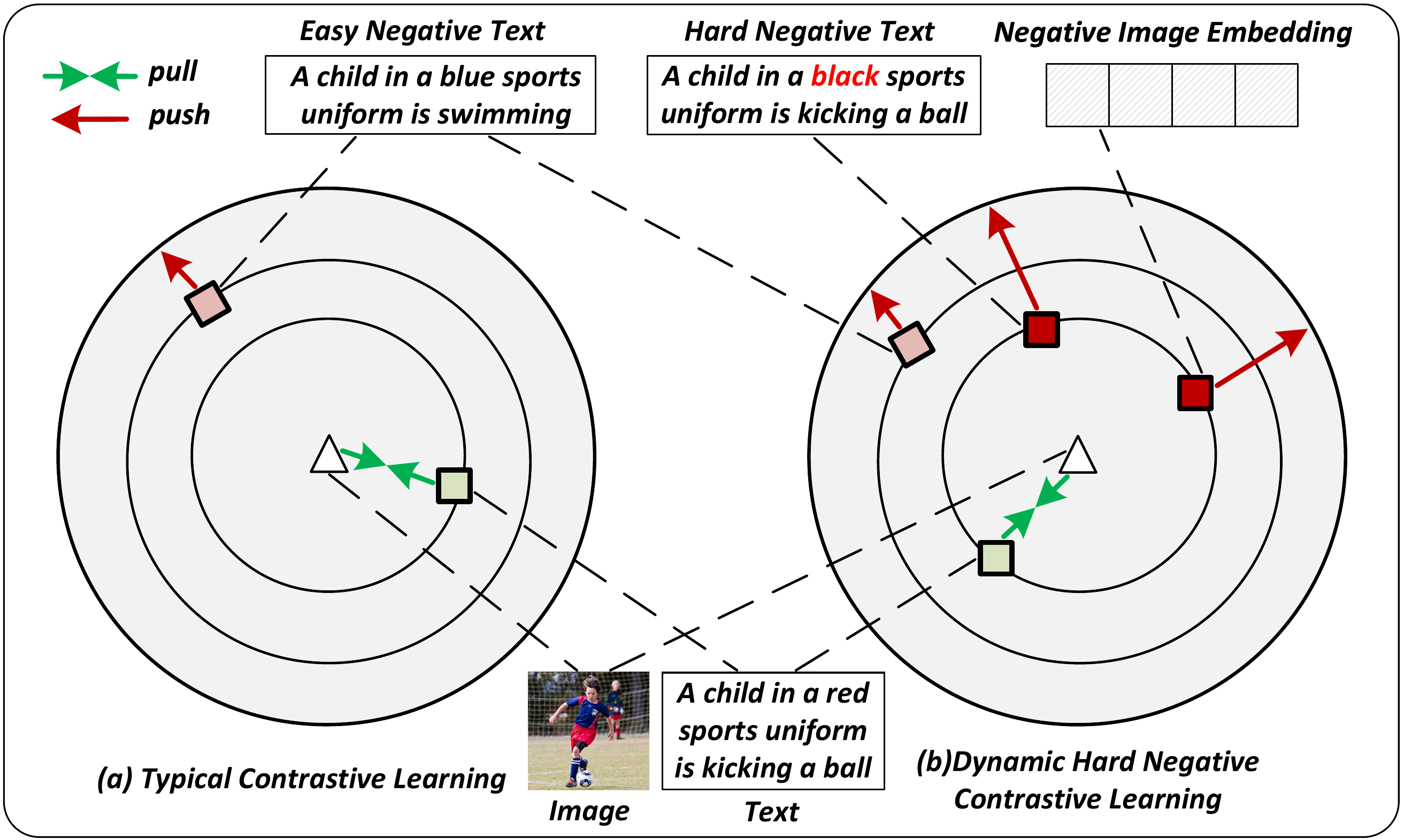}
    \caption{(a) The typical contrastive learning framework. (b) The proposed Dynamic Hard Negative Contrastive Learning, in which adjustments are specifically applied to the strategies for the distinction of hard negative samples and the alignment of positive samples. }
    \label{fig:3}
\end{figure}

To address these limitations, we propose dynamic hard negative contrastive learning (illustrated in Fig.~\ref{fig:3}), which dynamically adjusts its learning strategy based on sample difficulty, specifically enhancing the model's ability to distinguish difficult samples, and also establishes crucial semantic differences between hard negative samples and their corresponding positive samples in both visual and textual modalities, further improving the model's capacity to discern fine-grained semantic distinctions between them.

\paragraph{Contrastive Loss} \hspace*{0.2cm}Our method uses a contrastive loss, which is applied to a text and image pair \((T, I)\) as input and consists of two components: (i) an image encoder \(e_I = f_v(I)\); (ii) a text encoder \(e_T = f_t(T)\). In this setting, the text-to-image similarity score is computed as:
\begin{equation}
S(T, I) = \frac{e_T^T e_I}{\| e_T \| \| e_I \|} \label{eq:S_hat_base}
\end{equation}

As with most contemporary VLMs, we employ the contrastive loss as one of our losses for each batch:
\begin{multline} \label{eq:L_cont_multline_style}
L_{{cont}} = - \sum_i \Biggl( \log \left( \frac{\exp(S(T_i, I_i) / \tau)}{\sum_j \exp(S(T_i, I_j) / \tau)} \right) \\ 
+ \log \left( \frac{\exp(S(T_i, I_i) / \tau)}{\sum_k \exp(S(T_k, I_i) / \tau)} \right) \Biggr) 
\end{multline}
where \(\tau\) is a temperature parameter.

\paragraph{Multimodal Hard Negative Loss.}\hspace*{0.2cm} To enhance the model's ability to distinguish negative samples, we specially introduce a negative sample loss term. By simultaneously handling negative samples from both visual and textual perspectives, we help the model more accurately distinguish these challenging negative samples, thereby improving the effectiveness of contrastive learning.

The visual negative loss is designed to enhance the similarity differences between the visual negative embeddings \( I_n \) and the original image \( I\). Specifically, it is computed as:
\begin{equation}
L_{{neg}}^{{visual}} = \sum_{(I, T) \in B} -\log \left( \frac{1}{\sum_{I_n \in I_{hs}} \exp(S(I, I_n))} \right)
\end{equation}
where \(I_{hs}\) is the set of all such generated visual hard negative embeddings \(I_n\) for an original image \(I\), and \( S(I, I_n) \) represents the similarity between the original image \( I \) and each visual hard negative \( I_n \) in the set \( I_{hs} \). By addressing the subtle semantic differences between hard visual negatives, this loss encourages the model to better distinguish visually similar negative samples.

Next, the textual negative loss is defined. This loss enhances the similarity differences between the original text \( T \) and the hard negative texts \( T_n \). The textual negative loss is given as follows:
\begin{equation}
L_{{neg}}^{{textual}} = \sum_{(I, T) \in B} -\log \left( \frac{1}{\sum_{T_n \in T_{hs}} \exp(S(T, T_n))} \right)
\end{equation}
where \(T_{hs}\) is the set of all types of such textual hard negative samples \(T_n\) generated for an original text \(T\), and \( S(T, T_n) \) denotes the similarity between the original text \( T \) and each textual hard negative \( T_n \). This loss encourages the model to distinguish the original text from semantically similar hard negative texts by minimizing their similarity.

Finally, the total negative loss is obtained by combining the visual and textual negative losses:
\begin{equation}
L_{{neg}} = L_{{neg}}^{{visual}} + L_{{neg}}^{{textual}}
\end{equation}

By integrating the losses for both textual and visual negative samples, this approach aims to comprehensively reduce the similarity between negative and positive samples across both modalities, thereby enhancing the model's ability to distinguish negative samples. This not only improves the model's performance in cross-modal alignment tasks but also boosts its robustness by focusing on difficult-to-distinguish negative samples in both image and text domains.

\begin{table*}[t]
    \centering
    \small
    \renewcommand{\arraystretch}{0.9} 
    \setlength{\tabcolsep}{3pt} 
    \resizebox{\linewidth}{!}{
    \begin{tabular}{l c c c c c c c c c c c c c c c}
        \toprule
        \multirow{3}{*}{\textbf{Model}} & \multirow{3}{*}{\textbf{\#Params}} & \multicolumn{3}{c}{\textbf{ARO}} & 
        \multicolumn{10}{c}{\textbf{VALSE}} \\
        \cmidrule(lr){3-5} \cmidrule(lr){6-15}

         &  & \multirow{2}{*}{\textbf{Relation}} & \multirow{2}{*}{\textbf{Attribute}} & \multirow{2}{*}{\textbf{Avg}} & 
         \multicolumn{1}{c}{\textbf{Existence}} & 
         \multicolumn{1}{c}{\textbf{Plurality}} & 
         \textbf{Counting} & 
         \multicolumn{1}{c}{\textbf{Sp.rel.}} & 
         \multicolumn{2}{@{\hspace{-0.5em}}c@{}}{\textbf{Actions}} & 
         \multicolumn{2}{c}{\textbf{Coreference}} & 
         \multirow{2}{*}{\textbf{Foil-it!}} & 
         \multirow{2}{*}{\textbf{Avg}} \\

         &  &  &  &  &  
         \multicolumn{1}{c}{quantifiers} & 
         \multicolumn{1}{c}{number} & 
         & 
         \multicolumn{1}{c}{relations} & 
         \multicolumn{1}{c}{repl.} & 
         \multicolumn{1}{c}{actant swap} & 
         \multicolumn{1}{c}{standard} & 
         \multicolumn{1}{c}{clean} & 
         & \\
        
        \midrule
        
        BLIP \cite{BLIP} & 583M & 59.0 & 88.0 & 73.5 & 86.3 & 73.2 & 68.1 & 71.5 & 77.2 & 61.1 & 53.8 & 48.2 & 93.8 & 70.0 \\
        BEIT3 \cite{BEiT} & 1.9B & 60.6 & 74.6 & 67.6 & 77.4 & 74.6 & 68.8 & 74.0 & 86.7 & 65.2 & 50.0 & 44.2 & 96.0 & 70.4 \\
        BLIP2 \cite{BLIP-2} & 3.4B & 41.2 & 71.3 & 56.3 & 55.5 & 71.5 & 66.0 & 62.4 & 83.6 & 51.6 & 48.6 & 51.9 & 95.9 & 65.4 \\
        MiniGPT-4 \cite{MiniGPT-4} & \textgreater 9B & 46.9 & 55.7 & 51.3 & 65.5 & 72.5 & 67.4 & 68.4 & 83.2 & 58.8 & 52.6 & 51.0 & 95.8 & 68.4 \\
        \midrule
        \multicolumn{15}{l}{\textbf{\textit{Scene Graph relied method}}} \\
        syn-CLIP \cite{syn-CLIP} & 151M & 71.4 & 66.9 & 69.2 & - & - & - & - & - & - & - & - & - & - \\
        \midrule
        \multicolumn{15}{l}{\textbf{\textit{Segmentation \& LLM relied method}}} \\
        DAC-LLM \cite{DAC} & 151M & 81.3 & 73.9 & 77.6 & - & - & - & - & - & - & - & - & - & - \\
        DAC-SAM \cite{DAC} & 151M & 77.2 & 70.5 & 73.9 & - & - & - & - & - & - & - & - & - & - \\
        \midrule
        \multicolumn{15}{l}{\textbf{\textit{Hard Negative based method}}} \\
        
        CLIP \cite{CLIP} & 151M & 59.3 & 62.9 & 61.1 & 68.7 & 57.1 & 61.0 & 65.4 & 77.8 & 71.8 & 54.1 & 51.0 & 89.8 & 65.3 \\
        CyCLIP \cite{CyCLIP} & 151M & 59.1 & 65.4 & 62.3 & 69.3 & 58.3 & 61.0 & 66.4 & 78.1 & 72.0 & 53.2 & 51.6 & 88.8 & 65.5 \\
        SDS-CLIP \cite{SDS-CLIP} & 151M & 53.0 & 62.0 & 57.5 & - & - & - & - & - & - & - & - & - & - \\
        NegCLIP \cite{Neg-CLIP} & 151M & 80.2 & 70.5 & 75.4 & 76.8 & 71.7 & 65.0 & 72.9 & 81.6 & 84.7 & 58.6 & 53.8 & 91.9 & 71.6 \\
        CLIP-SVLC \cite{SVLC} & 151M & 80.6 & 73.0 & 76.8 & - & - & - & - & - & - & - & - & - & - \\
        
        CE-CLIP \cite{CE-CLIP} & 151M & 83.0 & 76.4 & 79.7 & 78.6 & 77.7 & 64.4 & 74.4 & 81.2 & 88.6 & 54.7 & 54.8 & 93.7 & 72.5 \\
        
         \rowcolor{gray!20}\textbf{Ours} & 151M & \textbf{83.8} & \textbf{77.0} & \textbf{80.4} & \textbf{84.0} & \textbf{78.7} & \textbf{64.9} & \textbf{76.3} & \textbf{81.9} & \textbf{88.1} & \textbf{56.0} & \textbf{58.6} & \textbf{94.4} & \textbf{75.9} \\
        \bottomrule
    \end{tabular}
    }
    \caption{Results (\%) on \textbf{ARO} and \textbf{VALSE}. Our proposed AHNPL achieves new state-of-the-art (SOTA) results on the VALSE benchmark.}
    \label{tab:1}
\end{table*}

\paragraph{Dynamic Margin Contrastive Loss.}\hspace*{0.2cm}In contrastive learning, the alignment of positive samples and the distinction of negative samples are key to improving model performance. Existing methods typically use a fixed margin (Margin Loss) threshold to adjust the similarity between positive and negative samples, but they still perform poorly when handling samples of varying difficulty.

When it comes to positive samples, such fixed margin approaches actually hold the assumption that all samples have the same level of difficulty. As a result, when handling challenging samples, the model tends to treat them the same as simple samples, making it difficult to capture the subtle semantic relationships, ultimately reducing the model's ability to align positive samples effectively. For negative samples, these methods generally introduce them into the contrastive learning framework without considering the variations in their difficulty. Although hard negatives are semantically similar to positive samples, they are characterized by subtle semantic differences. This subtle semantic difference makes it challenging for the model to effectively distinguish between positive samples and hard negatives, thereby limiting the model’s performance.

To effectively enhance the model's capability in aligning hard positive samples, we introduce a learnable parameter \(a\), which is initialized as a random value from a standard normal distribution. We set a lower bound of 0.2 for \(a\) to ensure it maintains a positive value throughout the entire training process. When the similarity between positive samples is low or there are significant semantic differences, the model automatically increases its focus on these challenging positive pairs, thus improving alignment accuracy.

Specifically, we compare the similarity \(S(I, T)\) of the positive sample pairs with the learnable parameter \(a\) to adjust the alignment strategy for positive samples:
\begin{equation}
L_{{mar}}^{+} = \sum_{(I, T) \in B} \max(0, a - S(I, T))
\end{equation}

This loss function ensures that positive samples with greater difficulty receive more attention during training, enabling the model to learn more effectively from challenging positive samples, thereby improving its overall performance in aligning positive samples.

To enable the model to better distinguish difficult negative samples, we introduce an adaptive threshold updating mechanism in margin loss based on the difficulty of the samples. The specific loss function is defined as follows:
\begin{equation}
L_{{mar}}^{-} = \sum_{(I, T) \in B} \sum_{T_n \in T_{hs}} \max(0, S(I, T_n) - S(I, T) + M_n^t)
\end{equation}
where \(M_n^t\) is the adaptive threshold computed for each textual hard negative \(T_n\) in the training step \(t\). This threshold is computed as follows:

\begin{equation}
M_n^{t} = \frac{1}{|B|} \sum_{(I, T) \in B} \left( S^{t-1}(I, T) - S^{t-1}(I, T_n) \right)
\end{equation}
where \( B \) represents the set of samples in the current training batch, \( S^{t-1}(I, T) \) and \( S^{t-1}(I, T_n) \) represent the similarity between image \( I \) and the positive text \( T \), as well as the similarity between the image \( I \) and a textual hard negative \( T_n \), respectively, in the previous training step.

The final margin loss combining both the negative margin loss and the positive margin loss is defined as:
\begin{equation}
L_{{mar}} = L_{{mar}}^{+} + L_{{mar}}^{-}
\end{equation}

This adaptive threshold updating strategy allows the model to dynamically adjust its learning strategy during the training process, ensuring effective alignment of positive samples and proper distinction of negative samples.

The final loss function can be expressed as follows:
\begin{equation}
L_{{total}} = L_{{cont}} + L_{{neg}} + L_{{mar}}
\end{equation}

\section{Experiments}
\subsection{Experimental Settings}
All experiments are conducted using the PyTorch framework on a single NVIDIA A40 GPU and an Intel\textsuperscript{\textregistered} Xeon\textsuperscript{\textregistered} Gold 6330 CPU, running on Ubuntu 22.10. During the training phase, we initialize a pretrained CLIP model and fine-tune it on the MSCOCO \cite{COCO} dataset for 10 epochs with a batch size of 128. The learning rate is set to \(2 \times 10^{-5}\), and weight decay is set to 0.1. 

\begin{table}[t]
    \centering
    \scriptsize 
    \renewcommand{\arraystretch}{0.9} 
    \setlength{\tabcolsep}{2pt} 
    \newcolumntype{C}{>{\centering\arraybackslash}X}
    \begin{tabularx}{\columnwidth}{l *{10}{C}}
        \toprule
        \multirow{2}{*}{\textbf{Model}} & \multicolumn{4}{c}{\textbf{REPLACE}} & \multicolumn{3}{c}{\textbf{SWAP}} & \multicolumn{3}{c}{\textbf{ADD}} \\
        \cmidrule(lr){2-5} \cmidrule(lr){6-8} \cmidrule(lr){9-11}
                         & Obj. & Att. & Rel. & Avg. & Obj. & Att. & Avg. & Obj. & Att. & Avg. \\
        \midrule
        Human    & 100 & 99.0 & 97.0 & 98.7 & 99.0 & 100 & 99.5 & 99.0 & 99.0 & 99.0 \\
        \midrule
        Vera \cite{vera}    & 49.4 & 49.6 & 49.1 & 49.4 & 49.4 & 49.2 & 49.3 & 49.4 & 49.6 & 49.5 \\
        Grammar \cite{Grammar}   & 50.0 & 50.0 & 50.0 & 50.0 & 50.0 & 50.0 & 50.0 & 50.0 & 50.0 & 50.0 \\
        BLIP2    & - & - & - & 86.7 & - & - & 69.8 & - & - & 86.5 \\
        \midrule
        CLIP              & 90.9 & 80.0 & 69.2 & 80.2 & 61.4 & 64.0 & 62.7 & 77.2 & 68.2 & 72.7 \\
        NegCLIP           & 92.7 & 85.9 & 76.5 & 85.0 & 75.2 & 75.4 & 75.3 & 88.8 & 82.8 & 85.8 \\
        CE-CLIP           & 93.1 & 88.8 & 79.0 & 87.0 & 72.8 & 77.0 & 74.9 & 92.4 & 93.4 & 92.9 \\
        
         \rowcolor{gray!20}\textbf{Ours}          & \textbf{93.2} & \textbf{88.9} & \textbf{80.1} & \textbf{87.4} & \textbf{76.3} & \textbf{75.8} & \textbf{76.1} & \textbf{97.2} & \textbf{94.5} & \textbf{95.9} \\
        \bottomrule
    \end{tabularx}
    
    \caption{Results (\%) on \textbf{SugarCrepe}. Vera and Grammar are text-only models.}
    \label{tab:2}
\end{table}

\subsection{Dataset}
\paragraph{Training Datasets.} The CLIP model is originally pretrained on 400 million image-text pairs sourced from the web, and we continue to fine-tune the model based on this pretraining. In our experiments, we use the widely adopted cross-modal text-image retrieval dataset MSCOCO \cite{COCO}. MSCOCO is chosen for its rich annotations of diverse objects, attributes, and relationships, which enhance the model's semantic understanding in multimodal tasks.

\paragraph{Evaluation Dataset.} We evaluate our method on several vision-language compositional benchmarks: ARO, VALSE \cite{VALSE}, and SugarCrepe \cite{SugarCrepe} (a bias-mitigated version of CREPE \cite{crepe}). Each test example in these datasets includes an image along with a corresponding correct description and a modified incorrect description. The model’s task is to determine which description is correct.
\begin{itemize}
    \item \textbf{ARO} \cite{Neg-CLIP}. This is a CR benchmark that includes positive or negative captions and evaluates sensitivity to word order. Word-order negative sentences are created by reordering words, which changes sentence semantics in attributes, relationships, and word order meaning. 
    \item \textbf{VALSE} \cite{VALSE}. This benchmark evaluates VLMs' understanding of linguistic phenomena. It includes six tests on structures: morphological syntax, verb-argument structure, and word order. These tests require models to accurately link visual elements to linguistic descriptions, assessing visual understanding and alignment of these phenomena.
    \item \textbf{SugarCrepe} \cite{SugarCrepe}. This benchmark uses LLMs to generate fluent and meaningful negative samples, avoiding biases introduced by traditionally used rule-based templates. It employs an adversarial improvement mechanism to minimize evaluation biases and ensures the reliability of results.
\end{itemize}

\subsection{Overall Results}
We present the evaluation results for the ARO and VALSE benchmarks in Tab.~\ref{tab:1}, and Tab.~\ref{tab:2} shows the evaluation results on the SugarCrepe benchmark. As a widely used evaluation benchmark in this field, VALSE provides a solid basis for comparing our model against various mainstream models. Our AHNPL demonstrates effective improvements over all methods that utilize hard negatives, achieving excellent results across all benchmarks.

\begin{table}[t]
    \centering
    \scriptsize
    \renewcommand{\arraystretch}{0.9} 
    \setlength{\tabcolsep}{2pt}
    \begin{tabular}{lcccccccc}
        \toprule
        \textbf{Model} & \textit{negatives} & \textit{MHNL} & \textit{DMCL} & \textbf{ARO-R} & \textbf{ARO-A} & \textbf{VALSE} & \textbf{SugarCrepe} & \textbf{Avg} \\
        \midrule
        CLIP &  &  &  & 59.3 & 62.9 & 67.0 & 71.9 & 65.3 \\
     
        & \checkmark &  &  & 81.6 & 72.0 & 74.2 & 80.3 & 77.0 \\
        & \checkmark & \checkmark &  & 82.1 & 72.9 & 73.0 & 85.1 & 78.3 \\
        & \checkmark &  & \checkmark & 80.1 & 71.9 & 72.7 & 85.5 & 77.6 \\
         \rowcolor{gray!20}\textbf{Ours} & \checkmark & \checkmark & \checkmark & \textbf{83.8} & \textbf{77.0} & \textbf{75.9} & \textbf{86.5} & \textbf{80.8} \\
        \bottomrule
    \end{tabular}
    \caption{Ablation of losses, where \textit{negatives} represent image-text contrastive with additional hard negatives.}
    \label{tab:3}
\end{table}

As shown in Tab.~\ref{tab:1}, our method achieves new state-of-the-art (SOTA) results on the VALSE dataset, with an average matching score of 75.9\%, outperforming the current SOTA model CE-CLIP by 3.4\%. Other models, such as NegCLIP, also perform well with a score of 71.6\%, but our model still leads by 4.3\%, particularly excelling in the quantifiers and clean subtasks, where it outperforms CE-CLIP by 5.4\% and 3.8\%, respectively. On the ARO benchmark, AHNPL outperforms CE-CLIP in the ARO-Relation task, indicating an improvement in relationship understanding. In the ARO-Attribute task, AHNPL slightly surpasses CE-CLIP, suggesting that our model is comparable to the current SOTA model in attribute understanding.

SugarCrepe is a debiased dataset that effectively avoids the biases introduced by traditional rule-based negative sample generation, making it a benchmark of relatively high difficulty. We choose only the best-performing and most representative models for evaluation on this benchmark. On the SugarCrepe benchmark, AHNPL achieves higher overall average scores than CE-CLIP, particularly in the SWAP and ADD tasks for the Object category, where it outperforms CE-CLIP by 3.5\% and 4.8\%, respectively. These results further demonstrate that our model has effective advantages in handling complex semantic relationships and reducing language bias.

\subsection{Ablation Study}
We conduct ablation experiments on multiple enhanced versions of the CLIP base model to evaluate the contribution and effectiveness of each component in our method. The specific impact of different loss functions on model performance is shown in Tab.~\ref{tab:3}. First, the results indicate that the introduction of hard negative samples effectively improves the model's performance, with an improvement of up to 11.4\% compared to the zero-shot CLIP, highlighting their key role in contrastive learning. Furthermore, compared to merely introducing negative samples into contrastive learning, each individual loss function we introduce shows varying degrees of performance improvement across all benchmarks. When all the loss functions are combined, the model achieves optimal performance. This improvement is fully reflected in our final model, demonstrating the superiority of the method in extending and enhancing contrastive learning objectives.

\begin{table*}[ht] 
    \centering
    \includegraphics[width=\textwidth]{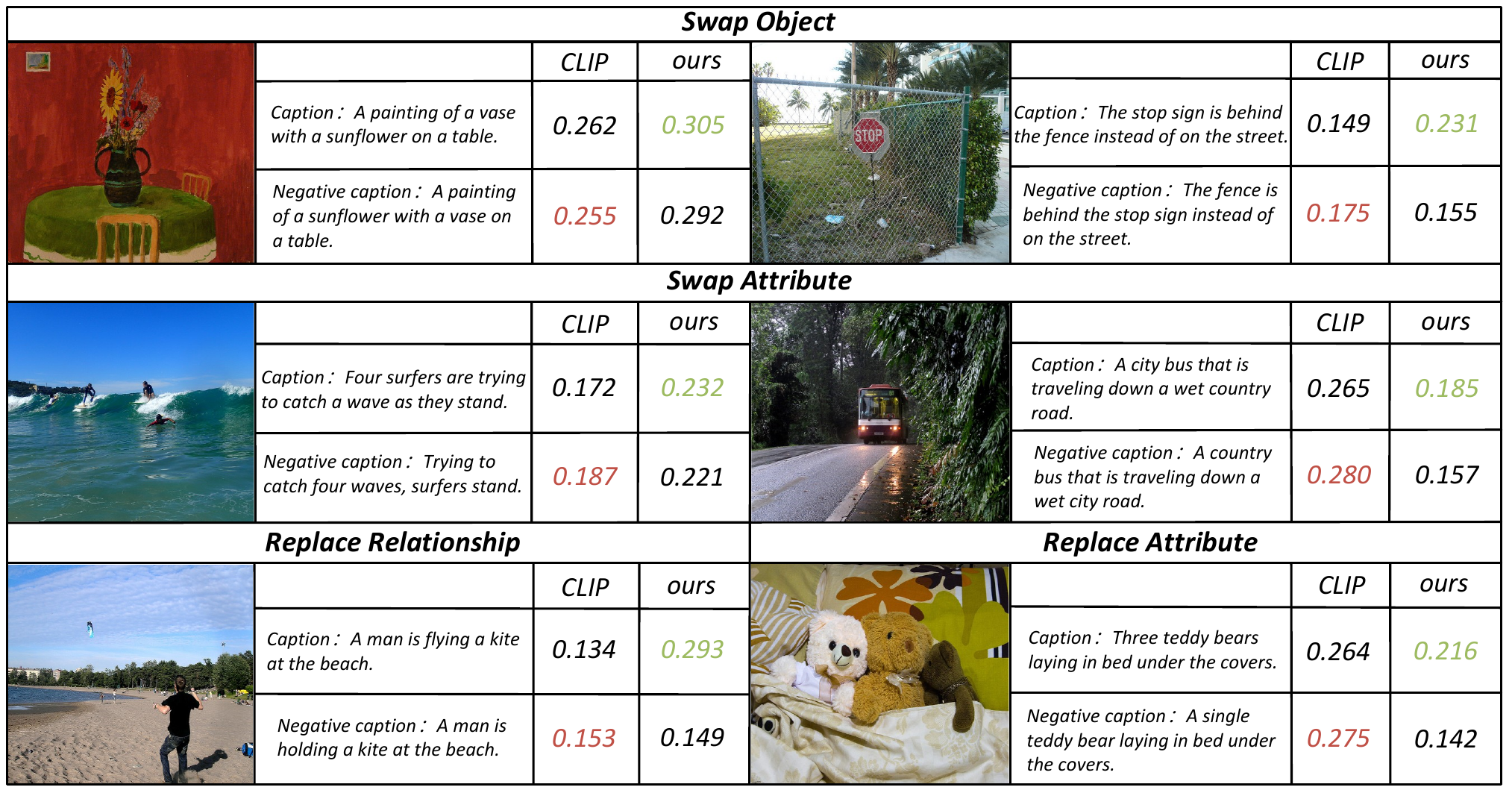} 
    \caption{Predictions of CLIP and AHNPL on SugarCrepe tasks: Swap Object, Swap Attribute, Replace Relationship and Replace Attribute. The score represents the similarity score between the caption and the corresponding image as assessed by CLIP/AHNPL. The model selects the caption with the higher similarity score as the correct one.}
    \label{tab:4}
\end{table*}

\subsection{Case Study}
We present several case studies illustrating the performance of CLIP and AHNPL across four subtasks of the SugarCrepe benchmark, as shown in Tab.~\ref{tab:4}. SugarCrepe uses LLMs to generate fluent captions with common sense, thereby challenging VLMs to distinguish negative captions effectively. In the “Swap Object” task, where models must understand object relationships (e.g., “A painting of a vase with a sunflower on a table” vs. “A painting of a sunflower with a vase on a table”), AHNPL outperforms CLIP. In the “Swap Attribute” task requiring accurate identification of object attributes (e.g., “Four surfers are trying to catch a wave as they stand” vs. “Trying to catch four waves, surfers stand”), CLIP struggles, whereas AHNPL consistently selects the correct caption. For the “Replace Relationship” and “Replace Attribute” tasks, which involve subtle distinctions (e.g., “A man is flying a kite at the beach” vs. “A man is holding a kite at the beach” or “Three teddy bears laying in bed under the covers” vs. “A single teddy bear laying in bed under the covers”), AHNPL effectively handles these nuances through negative caption contrastive learning, outperforming CLIP.

\subsection{Visualization}
To demonstrate the clear effectiveness of our proposed visual perturbation strategy, a compelling visualization study is provided. Tab.~\ref{tab:5} shows two illustrative examples, where each example generates textual negative samples using the methods in Section~3.1. Specifically, for each example, we extract \( e_{I_{orig}} \), \( e_{T_{orig}} \) and \( e_{T_{neg}} \) using zero-shot CLIP, AHNPL at epoch 5 and 10 (training end), respectively. The visual negative embedding \( e_{I_{neg}} \) is then computed from these extracted embeddings via our proposed visual perturbation. We illustrate the cosine distances between key pairs of these four embeddings. The changes in cosine distances demonstrate that the visual perturbation strategy effectively pushes the original image features towards negative texts, improving feature distinguishability and thereby helping the model capture subtle positive-negative semantic differences.

\begin{table}
    \centering
    \includegraphics[width=1\linewidth]{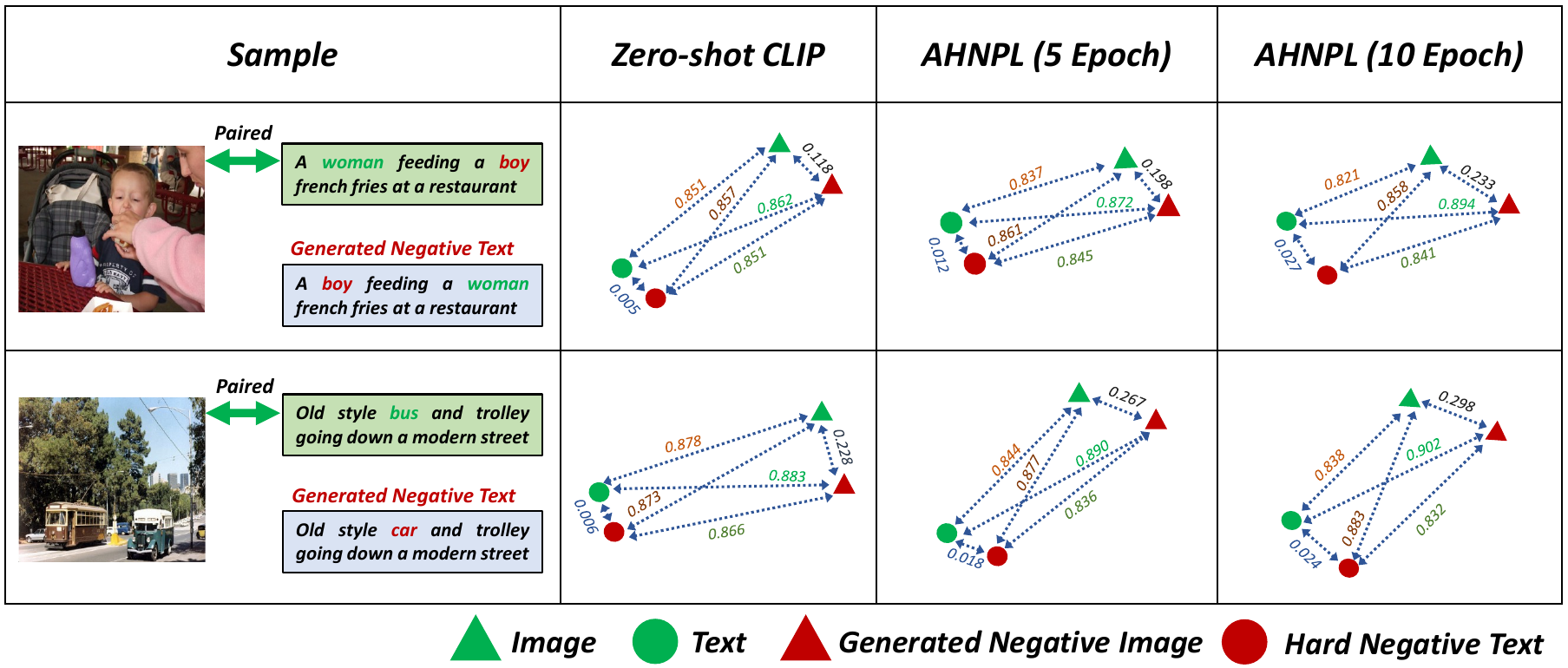}
    \caption{Impact of AHNPL training on the embedding space distances of two illustrative examples. Numerical values represent cosine distances between key embedding pairs.}
    \label{tab:5}
\end{table}

\section{Conclusion}
This paper proposes AHNPL, enhancing VLMs' performance in CR tasks via Visual Perturbation with Semantic Shift and Dynamic Hard Negative Contrastive Learning. AHNPL transforms text-based hard negatives into semantically perturbed image negatives, effectively improving the overall performance. Additionally, Dynamic Hard Negative Contrastive Learning strengthens the model’s ability to align positive samples and distinguish hard negatives. Experimental results show that AHNPL outperforms existing methods in downstream tasks, showing robust complex semantic understanding in multimodal scenarios. Future work will explore integrating large-scale knowledge graphs and implicit knowledge to enhance complex semantic relationship acquisition.

\section*{Acknowledgments}
This work was supported by Humanities and Social Sciences Youth Foundation, Ministry of Education of the People's Republic of China (No. 24YJCZH135), and Science and Technology Projects of Henan Province (Nos. 242102211019 \& 252102210027).

\bibliographystyle{named}
\bibliography{ijcai25}

\end{document}